\documentclass[conference]{IEEEtran}

\usepackage{graphicx}
\usepackage{booktabs}
\usepackage[ruled]{algorithm2e}
\usepackage{xspace}
\usepackage{subcaption}
\usepackage{fixltx2e}
\usepackage{amsmath}
\usepackage{todonotes}
\usepackage{cite}
\usepackage[bookmarks=false]{hyperref}

\usepackage[usestackEOL]{stackengine}
\stackMath

\graphicspath{{figures/}}
\newcommand{\eg}{\emph{e.g.}}

\newcommand\specparen[2]{%
  \def\Krn{\kern2.5ex}%
  \def\useanchorwidth{T}%
  \setbox0=\hbox{\Krn\stackengine{0pt}{\scriptstyle#1}{\scriptstyle#2}{O}{c}{F}{F}{S}}%
  \stackon[2pt]{\stackunder[2pt]{)}{\makebox[\wd0][l]{\Krn$\scriptstyle#1$}}}%
                                   {\makebox[\wd0][l]{\Krn$\scriptstyle#2$}}%
}

\begin{document}

\title{Incomplete Dot Products for Dynamic Computation Scaling in Neural Network Inference}

\author{
 \IEEEauthorblockN{Bradley McDanel}
 \IEEEauthorblockA{Harvard University\\
 Cambridge, MA, USA\\
 Email: mcdanel@fas.harvard.edu}
 \and
 \IEEEauthorblockN{Surat Teerapittayanon}
 \IEEEauthorblockA{Harvard University\\
 Cambridge, MA, USA\\
 Email: steerapi@seas.harvard.edu}
 \and
 \IEEEauthorblockN{H.T. Kung}
 \IEEEauthorblockA{Harvard University\\
 Cambridge, MA, USA\\
 Email: kung@harvard.edu}
}

\maketitle

\begin{abstract}
We propose the use of incomplete dot products (IDP) to dynamically adjust the number of input channels used in each layer of a convolutional neural network during feedforward inference. IDP adds monotonically non-increasing coefficients, referred to as a "profile", to the channels during training. The profile orders the contribution of each channel in non-increasing order. At inference time, the number of channels used can be dynamically adjusted to trade off accuracy for lowered power consumption and reduced latency by selecting only a beginning subset of channels. This approach allows for a single network to dynamically scale over a computation range, as opposed to training and deploying multiple networks to support different levels of computation scaling. Additionally, we extend the notion to multiple profiles, each optimized for some specific range of computation scaling. We present experiments on the computation and accuracy trade-offs of IDP for popular image classification models and datasets. We demonstrate that, for MNIST and CIFAR-10, IDP reduces computation significantly, e.g., by $75\%$, without significantly compromising accuracy. We argue that IDP provides a convenient and effective means for devices to lower computation costs dynamically to reflect the current computation budget of the system. For example, VGG-16 with $50\%$ IDP (using only the first $50\%$ of channels) achieves $70\%$ in accuracy on the CIFAR-10 dataset compared to the standard network which achieves only $35\%$ accuracy when using the reduced channel set.
\end{abstract}

\IEEEpeerreviewmaketitle

\section{Introduction}
Inference with deep Convolutional Neural Networks (CNNs) on end or edge devices has received increasing attention as more applications begin to use sensor data as input for models running directly on the device (see, e.g., \cite{McMahanMRA16}). However, each trained CNN model has a fixed accuracy, size, and latency profile determined by the number of layers and parameters in each layer. The static nature of these models can be problematic in dynamic contexts, such as when running on mobile device, where the power and latency requirements of CNN inference may change based on the current battery life or computation latency allowance of the device. 

One approach to address these dynamic contexts is to train multiple CNN models of ranging sizes, such as by varying the number of parameters in each layer as in MobileNet~\cite{howard2017mobilenets}, and then selecting the appropriate model based on the current system requirements. However, this kind of approach requires storing multiple models of different sizes to support the desired computation flexibilities of the system. Ideally, we would instead train a single CNN that could dynamically scale across a computation range in high resolution, trading off accuracy for power consumption and latency as desired.

\begin{figure}
    \centering
    \includegraphics[width=\linewidth]{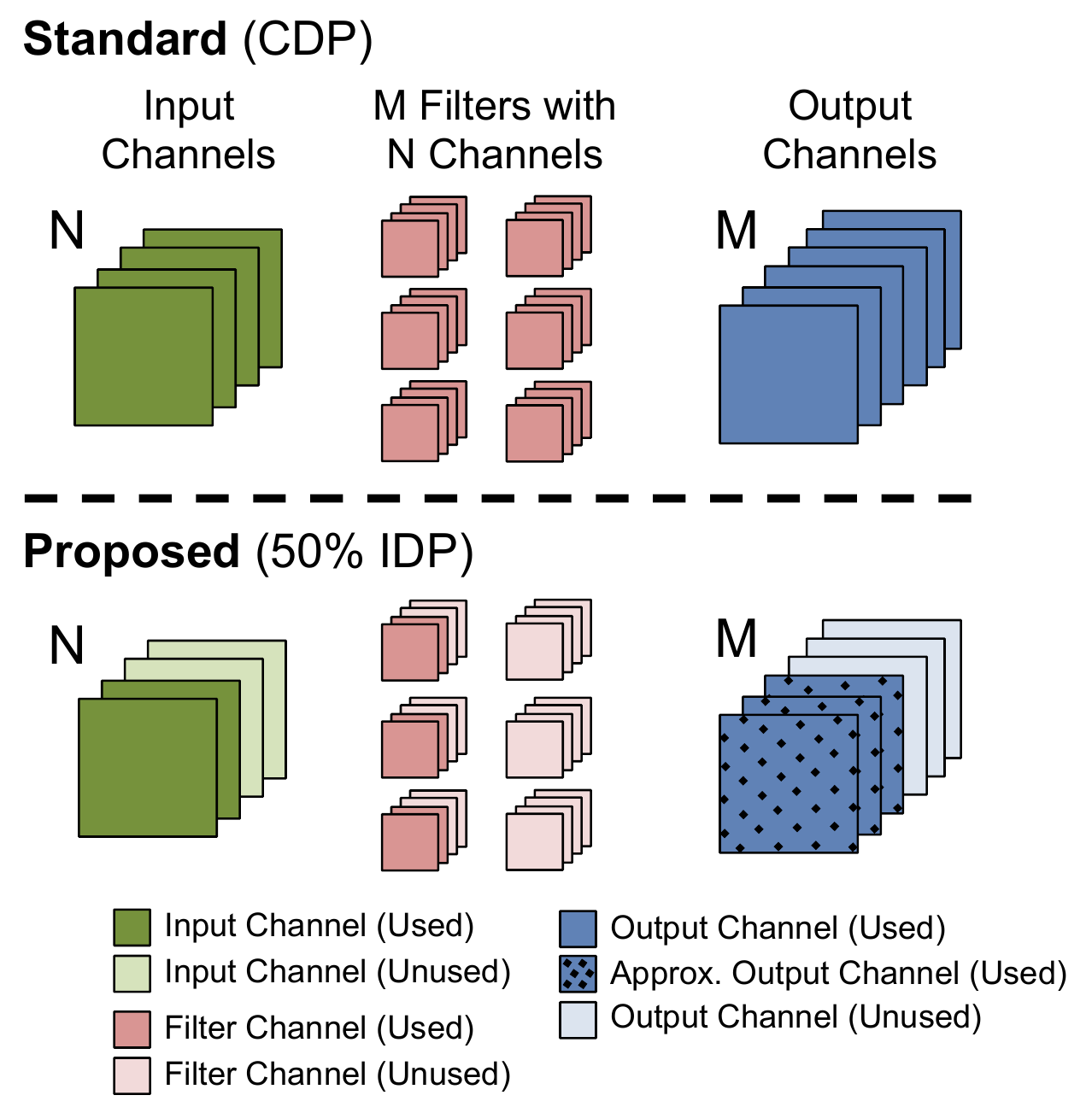}
    \caption{Contrasting computation using complete dot product (CDP) in standard networks and incomplete dot product (IDP) proposed in this paper for a convolutional layer. Under standard CNN, for each filter at a given layer, \textsf{N} input channels are used in the dot product computation (CDP), to compute the corresponding output channel. Under, for example, $50\%$ IDP, the filter uses the first $50\%$ of the input channels to compute the corresponding output channel, which is an approximation of the CDP. Furthermore, only 50\% of these filters are used since output channels for the other filters will not be utilized in the next layer. This leads to a $75\%$ reduction in computation for $50\%$ IDP.}
    \label{fig:savings-comparison}
\end{figure}

To provide a solution for dynamic scaling over a computation range, we  propose to modify CNN training, by adding a \textit{profile} of monotonically non-increasing channel coefficients for channels in a given layer. These coefficients provide an ordering for channels to allow channels with small coefficients to be dropped out during inference to save computation. By training a CNN with this profile, dot products performed during forward propagation may use only some of the beginning channels in each filter, without compromising accuracy significantly. We call such a dot product \textit{incomplete dot product} (IDP). As we show in our evaluation in Section~\ref{sec:eval}, using IDP on networks trained with a properly chosen profile achieves much higher accuracy relative to using IDP on standard CNNs (trained without channel coefficients). Using the IDP scheme, we are able to train a single CNN that can dynamically scale across a computation range.

Figure~\ref{fig:savings-comparison} contrasts forward propagation in a standard CNN layer using all input channels, which we refer to as \textit{complete dot product} (CDP), to that of using only half of the input channels ($50\%$ IDP). IDP has two sources of computation reduction: 1) each output channel is computed with a filter using only $50\%$ of the input channels (meaning the output is an approximation of using 100\% of channels) and 2) half of the filters are unused since their corresponding output channels are never consumed in the next layer. Therefore, $50\%$ IDP leads to a $75\%$ reduction in computation cost. In general, the IDP computation cost is $p^2$ times the original computation cost, where $p$ is the percentage of channels used in IDP. As we show in our analysis in Section~\ref{sec:eval}, IDP reduces computations (number of active input channels used in forward propagation) of several conventional network structures while still maintaining similar accuracy when all channels are used.

In the next section, we describe how IDP is integrated into multiple layer types (fully connected, convolution, and depthwise separable convolution) in order to train such networks where only a subset of channels may be used during forward propagation. We call such networks incomplete neural networks.

\section{Incomplete Neural Networks}
An incomplete neural network consists of one or more incomplete layers. An incomplete layer is similar to a standard neural network layer except that incomplete dot product (IDP) operations replace conventional complete dot product (CDP) operations during forward propagation. We begin this section by describing IDP during forward propagation and provide an explanation of how IDP is specifically added to fully-connected, convolutional, and depthwise separable convolutional layers.

\subsection{Incomplete Dot Product}
IDP adds a profile consisting of monotonically non-increasing coefficients $\boldsymbol{\gamma}$ to the components of the dot product computations of forward propagation. These coefficients order the importance of the components in decreasing order from the most to least important. Mathematically, the incomplete dot product (IDP) of two vectors $(a_1, a_2, \ldots, a_N)^\mathsf{T}$ and $(b_1, b_2, \ldots, b_N)^\mathsf{T}$ is a truncated version of the following expression
$$
{\displaystyle \sum _{i=1}^{N}\gamma_i  a_{i}b_{i}=\gamma_1 a_{1}b_{1}+\gamma_2 a_{2}b_{2}+\cdots +\gamma_N a_{N}b_{N}}
$$
which keeps only some number of the beginning terms, where $\gamma_1, \gamma_2, \ldots, \gamma_N$ are the monotonically non-increasing profile coefficients.

To compute IDP with a target IDP percentage, the beginning components, starting with $\gamma_1a_{1}b_{1}$, are accumulated until the target percentage of components is reached. The contribution of the remaining components, with smaller coefficients, is ignored, making the dot product incomplete. This is mathematically equivalent to setting the unused coefficients to $0$.

\subsection{Choice of Profile}
\label{sec:coef-functions}
There are multiple ways to set the profile $\boldsymbol{\gamma} = \{\gamma_1, \gamma_2, \ldots, \gamma_i, \ldots, \gamma_N\}$ corresponding to various policies which may favor dynamic computation scaling for certain ranges at the expense of other regions. In Section~\ref{sec:coef-eval}, we discuss the performance implications of the various profiles. The profiles evaluated in this paper, shown in Figure~\ref{fig:coef-functions}, are:  

\begin{figure}
    \centering
    \includegraphics[width=\linewidth]{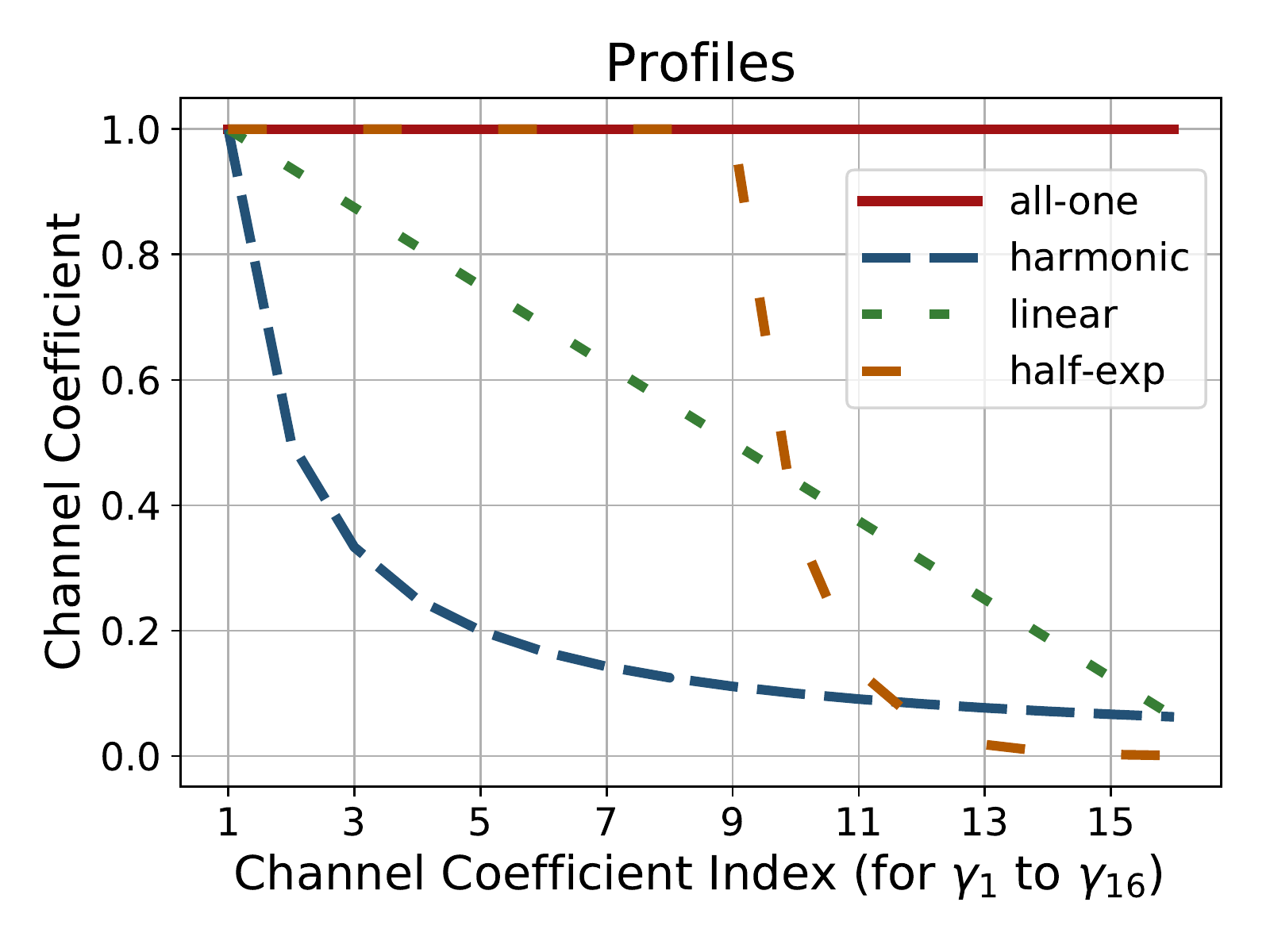}
    \caption{The profiles evaluated in this paper. Each channel coefficient index on the x-axis corresponds to an input channel (as shown in Figures~\ref{fig:incomplete} and \ref{fig:incomplete_separable}). The y-axis shows the value of each coefficient for a profile. The all-one (all coefficients equal $1$) profile corresponds to a standard network layer without channel coefficients. The other profiles provide different schemes for the coefficients. For instance, in half-exp, an exponential decay for second half of coefficients is used.}
    \label{fig:coef-functions}
\end{figure}

\subsubsection{all-one} corresponds to a standard network layer without a profile
$$
%\gamma_i1^N = 1
 %\gamma_n = 1
 %(\boldsymbol{\gamma})\mkern-10.5mu\mathop{\vphantom{)}}\limits_{n=1}^N = 1
% \textbf{all-one}(n) = 1
 \gamma_i = 1 \text{ for } i=1,\dots,N
$$
\subsubsection{harmonic} a harmonic decay 
$$
% \gamma_n
% \gamma_n(k) = \frac{k}{n}
 \gamma_i = \frac{1}{i}  \text{ for } i=1,\dots,N
% \textbf{harmonic}(n,k) = \frac{k}{n}
$$
\subsubsection{linear} a linear decay
$$
 \gamma_i = 1 - \frac{i}{N}  \text{ for } i=1,\dots,N
 %\gamma_n(k) = 1-\frac{k-1}{n}
% \textbf{linear}(n,k) = 1-\frac{k-1}{n}
$$
% \subsubsection{exponential} an exponential decay. Mathematically, we have
% $$
% \textbf{exponential}(n,k) = e^{-(k-1)n}
% $$
\subsubsection{half-exp} \textit{all-one} for the first half of terms and an exponential decay for latter half of terms
\[
  \gamma_i \ =
  \begin{cases}
    1,      & \text{if } i < \frac{N}{2}\\
    \exp{\left(\frac{N}{2} - i - 1 \right)}, & \text{otherwise}
  \end{cases}  \text{ for } i=1,\dots,N
\]

% \[
%   \textbf{half-exp}(n,k) = \left\{
%      \begin{array}{@{}l@{\thinspace}l}
%       1 &, \text{if } n < \frac{N}{2} \\
%       e^{-(k-1)(n-\frac{N}{2})} &, \text{if } n\ge\frac{N}{2}
%      \end{array},
%   \right.
% \]
% \subsubsection{step} a step function supporting multiple steps. Mathematically, we have
% \[
%   \textbf{step}(n, k, \{c_{i \in \{1,\dots,m\}}\}) = c_i, \text{if } \frac{(i-1)}{m}N < n \le \frac{i}{m}N
% \]
% where $N$ is the number of components, $n \in {1,\dots,N}$ and $k$ are the hyperpameters of each function.

% For fully connected layers, ...
% For depthwise separable convolutional layers, ...

\subsection{Incomplete Layers}
In this section, we describe how IDP is integrated into standard neural network layers, which we refer to as incomplete layers.

\subsubsection{Incomplete Fully Connected Layer}
A standard linear layer does the following computation:
$$y_{j} = \sum_{i=1}^{N}w_{ji}x_i,$$
where $j\in{\{1,\dots,M\}}$, $M$ is the number of output components, $N$ is the number of input components, $w_{ji}$ is the layer weight corresponding to the $j$-th output component and $i$-th input component, $x_i$ is the $i$-th input component and $y_j$ is $j$-th output component. An incomplete linear layer does the following computation instead:
$$y_{j} = \sum_{i=1}^{N}\gamma_i w_{ji}x_i,$$
where $\gamma_i$ is the $i$-th profile coefficient.

\subsubsection{Incomplete Convolution Layer}
\begin{figure}
    \centering
    \includegraphics[width=.73\linewidth]{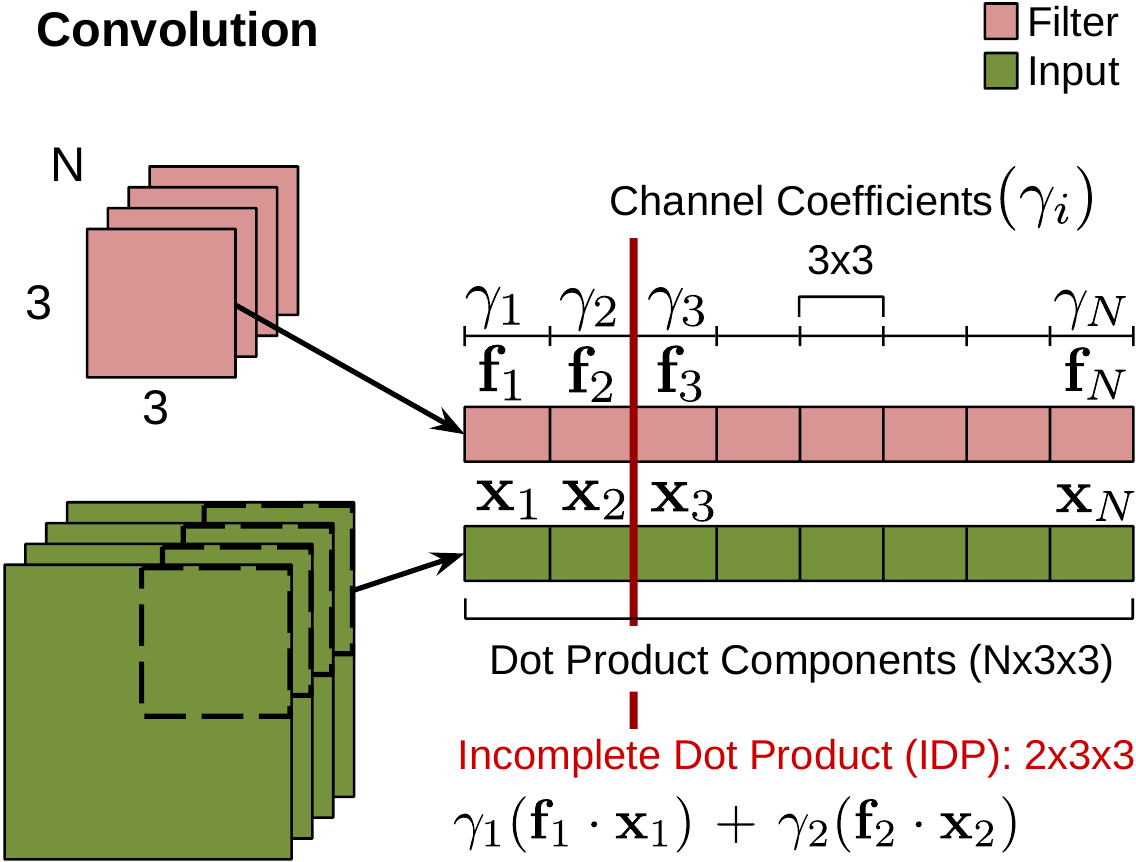}
    \caption{IDP computation for a filter $\mathbf{f}$ (green) consisting of $\mathsf{N}$ $3\times3$ kernels $\mathbf{f}_1, \mathbf{f}_2, \dots, \mathbf{f}_N$ and a patch of the input $\mathbf{x}$ (green) consisting of slices from $\mathsf{N}$ input channels, $\mathbf{x}_1, \mathbf{x}_2, \dots, \mathbf{x}_N$ which is highlighted by the dashed black lines. On the right, the filter and input are shown in vector form. The coefficients $\gamma_1, \dots, \gamma_N$ correspond to each input channel. The vertical dashed line (red) signifies an IDP where only the first two channels are used to compute the dot product $\gamma_1(\mathbf{f}_1 \cdot \mathbf{x}_1) + \gamma_2(\mathbf{f}_2 \cdot \mathbf{x}_2)$.}
    \label{fig:incomplete}
\end{figure}

A standard convolution layer does the following computation:
$$\mathbf{y}_{j} = \sum_{i=1}^{N}\mathbf{f}_{ji}*\mathbf{x}_i,$$
where $j\in{\{1,\dots,M\}}$, $M$ is the number of output channels, $N$ is the number of input channels, $\mathbf{f}_{ji}$ is the $i$-th channel of the $j$-th filter, 
%$\mathcal{F}_j = \{\mathbf{f}_i\}_{i\in\{1,\dots,N\}}$
$\mathbf{x}_i$ is the $i$-th input channel and $\mathbf{y}_j$ is the $j$-th output channel. When the input data is 2D, $\mathbf{f}_{ji}$ is a 2D kernel and $\mathbf{f}_{ji}*\mathbf{x}_i$ is a 2D convolution. For this paper, we use 2D input data in all experiments. An incomplete convolution layer does the following computation instead:
$$\mathbf{y}_{j} = \sum_{i=1}^{N}\gamma_i(\mathbf{f}_{ji}*\mathbf{x}_i),$$
where $\gamma_i$ is the profile coefficient for the $i$-th channel of a filter, as illustrated in Figure~\ref{fig:incomplete}.

% \subsubsection{Incomplete Depthwise Convolution Layer}
% A standard depthwise convolution layer does the following computation:
% $$\mathbf{y}_{i} = \mathbf{f}_i*\mathbf{x}_i,$$
% where $\mathbf{f}_i$ is the $i$-th depthwise kernel, $\mathbf{x}_i$ is the $i$-th input channel and $\mathbf{y}_i$ is the $i$-th output channel. An incomplete depthwise convolution layer does the following computation instead:
% $$\mathbf{y}_{i} = \gamma_i(\mathbf{f}_i*\mathbf{x}_i),$$ where $\gamma_i$ is the coefficient of $i$-th depthwise kernel.

\subsubsection{Incomplete Depthwise Separable Convolution Layer}
\begin{figure}
 \centering
 \includegraphics[width=\linewidth]{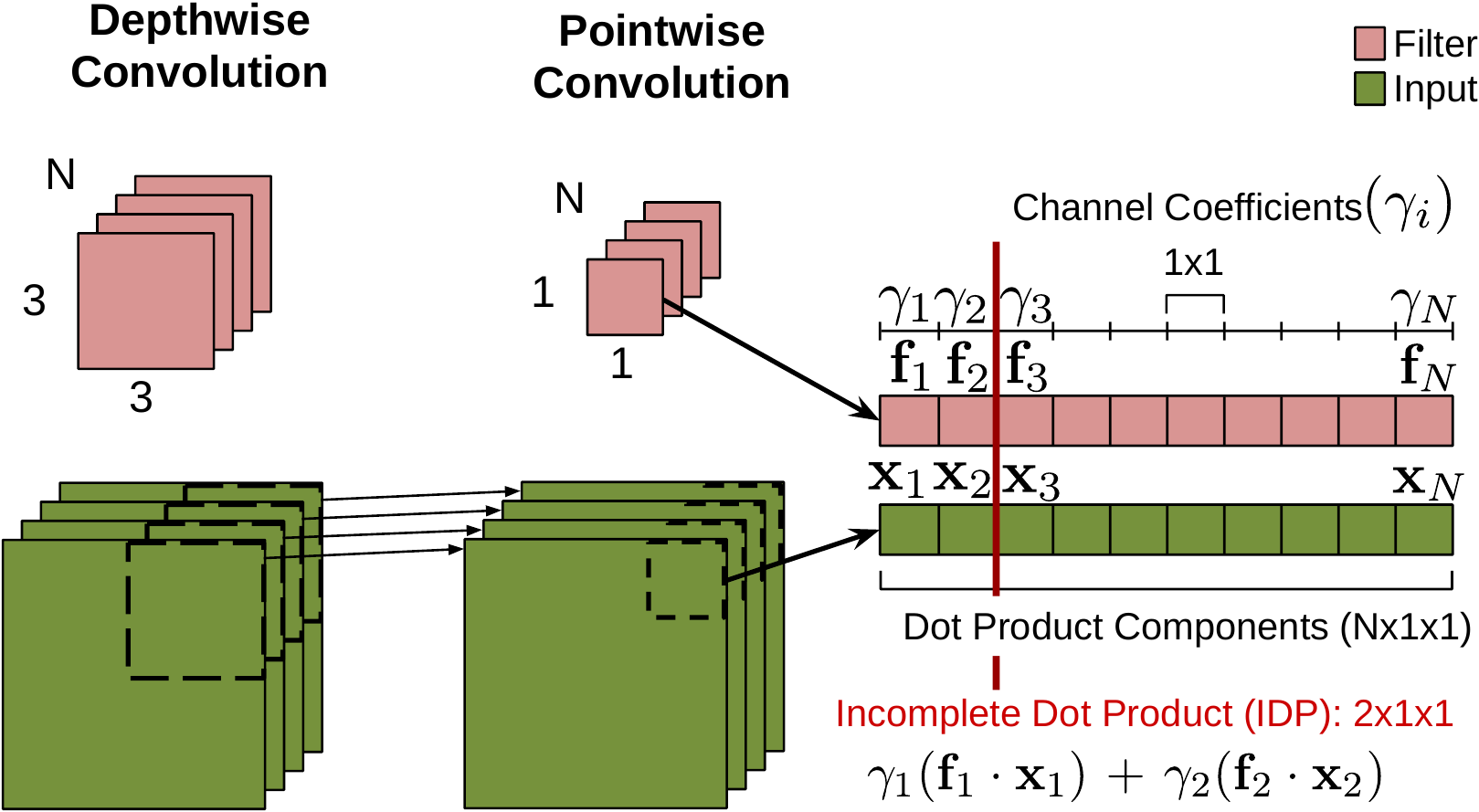}
 \caption{Incomplete depthwise separable convolution consists of depthwise convolution followed by pointwise convolution. In this illustration, IDP is only applied to the pointwise convolution. Note that IDP can also be applied to depthwise convolution such as illustrated in Figure~\ref{fig:incomplete}.}
 \label{fig:incomplete_separable}
\end{figure}

An incomplete depthwise separable convolution~\cite{howard2017mobilenets}~consists of depthwise convolution followed by pointwise convolution. To simplify the presentation in this work, IDP is only applied to the pointwise convolution. A standard depthwise separable convolution layer does the following computation:
$$\mathbf{y}_{j} = \sum_{i=1}^{N}\mathbf{g}_{ji}*(\mathbf{f}_{i}*\mathbf{x}_i),$$
where $j\in{\{1,\dots,M\}}$, $M$ is the number of output channels, $N$ is the number of input channels, $\mathbf{g}_{ji}$ is the $i$-th channel of the $j$-th pointwise filter, 
%$\mathcal{G}_j = \{\mathbf{g}_i\}_{i\in\{1,\dots,N\}}$
$\mathbf{f}_{i}$ is the $i$-th  channel of the depthwise filter, 
%$\mathcal{F}_j = \{\mathbf{f}_{ji}\}_{i\in\{1,\dots,N\}}$,
$\mathbf{x}_i$ is the $i$-th input channel and $\mathbf{y}_j$ is the $j$-th output channel. An incomplete depthwise convolution layer does the following computation instead:
$$\mathbf{y}_{j} = \sum_{i=1}^{N}\gamma_j(\mathbf{g}_{ji}*(\mathbf{f}_{i}*\mathbf{x}_i)),$$
where $\gamma_j$ is the profile coefficient of the $j$-th pointwise filter, as illustrated in Figure~\ref{fig:incomplete_separable}.

% \subsubsection{Binary layers}
% Each binary layer forces the layer weight values to be either -1 or 1.

\subsection{Incomplete Blocks}

An incomplete block consists of one or more incomplete layers, batch normalization and an activation functions. In this paper, Conv denotes an incomplete convolution block containing an incomplete convolution layer, batch normalization and an activation function. Similarly, S-Conv denotes an incomplete depthwise separable convolution block. FC denotes an incomplete fully connected block. B-Conv denotes an incomplete binary convolution block, where the layer weight values are either -1 or 1, and B-FC denotes an incomplete binary fully connected block.

\vspace{10mm}

\section{Multiple-Profile \\ Incomplete Neural Networks}
\label{sec:multi-profile}

\begin{figure}
    \centering
    \includegraphics[width=\linewidth]{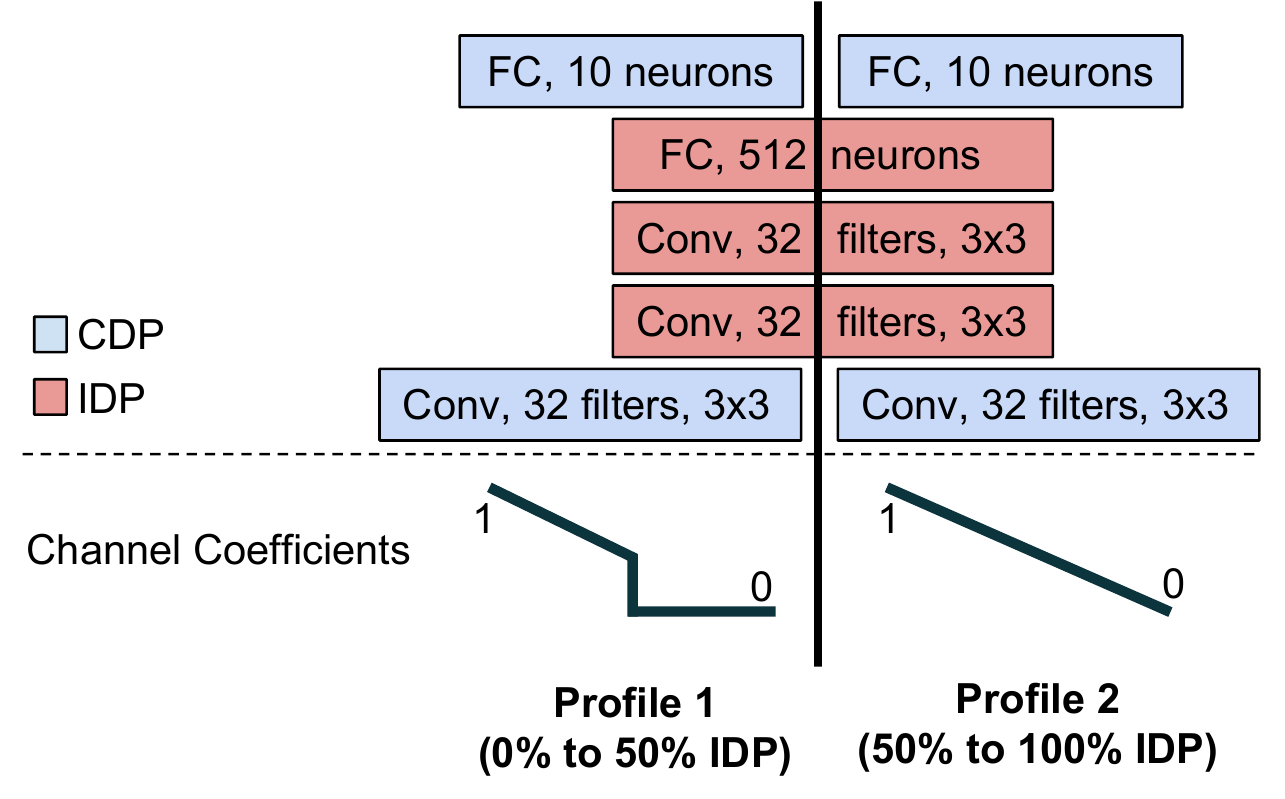}
    \caption{Training an incomplete neural network with two profiles. First, profile 1 is trained using only the first $50\%$ of channels in each filter for every IDP layer. Then profile 2 is trained using all channels in every IDP layer, but only updates the latter $50\%$ of channels in each filter. Each profile has an independent first and last layer. In this example, linear profiles (see Figure~\ref{fig:coef-functions}) are used.}
    \label{fig:training}
\end{figure}

Multiple profiles, each focusing on a specific IDP range, can be used in a single network to efficiently cover a larger computation range. Figure~\ref{fig:training} provides an overview of the training process for a network with two profiles (the process generalizes to three or more profiles). In the example, the first profile operates on the $0$-$50\%$ IDP range and the second profile operates on the $50$-$100\%$ IDP range. The training phase is performed in multiple stages, first training profile 1 and then fixing the learned weights and training profile 2. Specifically, when training profile 1, only the channels corresponding to the $0$-$50\%$ IDP range are learned. After training profile 1, profile 2 is trained in a similar manner, but only learns weights in the $50$-$100\%$ IDP range while still utilizing the previously learned weights of profile 1.  

At inference time, a profile is selected based on the current IDP percentage requirement. The IDP percentage is chosen by the application power management policy, which dictates computation scaling based on, \eg,~current battery budget of the device. For the two-profile example in Figure~\ref{fig:training}, when IDP is between $0\%$ and $50\%$, profile 1 is used, otherwise profile 2 is used.

While the middle IDP layers are shared across all profiles, each profile maintains a separate first and last layer. This design choice is optional. Experimentally, we found that maintaining a separate first and last layer helps improve the accuracy of each profile by adapting the profile to the subset of channels present in the IDP range. Generally, these separate layers add a small amount of memory overhead to the base model. For instance, for VGG-16~\cite{simonyan2014very} model used in our evaluation in Section~\ref{sec:eval}, the additional profile adds a 64, 3x3 filter convolutional layer (first layer) and a 10 neuron fully connected layer for the profile classifier (last layer). In this case, the second profile layers translate into a $3\%$ increase in total model size over a single profile VGG-16 network.

\section{Evaluation}
\label{sec:eval}
In order to show the general versatility of the approach, we incorporate IDP into several well studied models and evaluate on two datasets (MNIST~\cite{lecun1998mnist} and CIFAR-10~\cite{krizhevsky2014cifar}).

\subsection{Network Models}
\label{sec:networks}
\begin{figure*}
    \centering
    \includegraphics[width=0.8\linewidth]{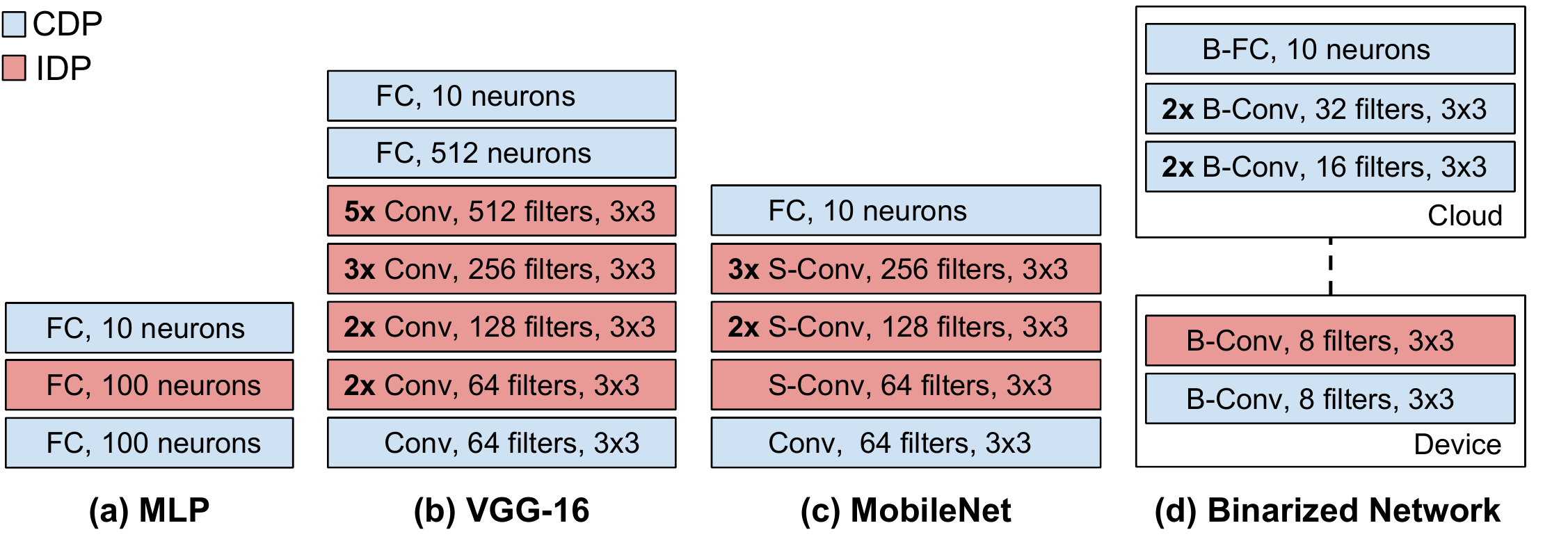}
    \caption{The networks used to evaluate IDP. The complete dot product (CDP) layers (blue) are standard convolutional layers. The IDP layers (red) are trained with a profile, described in Section~\ref{sec:coef-functions}, in order to use a subset of dot product components during inference across a dynamic computation scaling range. A multiplier (\eg,~\textbf{2x}) denotes multiple repeated layers of the same type. For MobileNet, each \textit{S-Conv} layer is a depthwise separable convolution layer as described in~\cite{howard2017mobilenets}. For the Binarized Network model, \textit{B-Conv} and \textit{B-FC} refer to a binary convolution layer and a  binary fully connected layer, respectively. For this model, we segment the layer between the device and the cloud, with the lower layers on device and higher layers in the cloud, as in~\cite{teerapittayanon2017distributed}. In the evaluation, when an IDP percentage is specified, such as $50\%$ IDP, all IDP layer shown will use this percentage of input channels during forward propagation.}
    \label{fig:networks}
\end{figure*}

The network models used to evaluate IDP are shown in Figure~\ref{fig:networks}. These networks cover several network types including MLPs, CNNs, Depthwise Separable CNNs (MobileNet~\cite{howard2017mobilenets}), and Binary Networks. 

\subsubsection{MLP}
This single hidden layer MLP model highlights the effects of IDP on a simple network. We use the MNIST dataset when evaluating this network structure.

\subsubsection{VGG-16}
VGG-16 is a large CNN; therefore it is of interest to see how IDP performs when a large number of input channels are not used at each layer. We use the CIFAR-10 dataset when evaluating this network structure.

\subsubsection{MobileNet}
This network structure is interesting because of the 1x1 convolution layer over the depth dimension which is computationally expensive when depth is large, making it a natural target for IDP. We use the CIFAR-10 dataset when evaluating this network structure.

\subsubsection{Binarized Networks}
Binarized Networks\cite{courbariaux2016binarized} are useful in low-power and embedded device applications. We combine these networks with a device and cloud layer segmentation as shown in Figure~\ref{fig:networks}. We use the MNIST dataset when evaluating this network structure.

\subsection{Impact of Profile Selection}
\begin{figure}
    \centering
    \includegraphics[width=\linewidth]{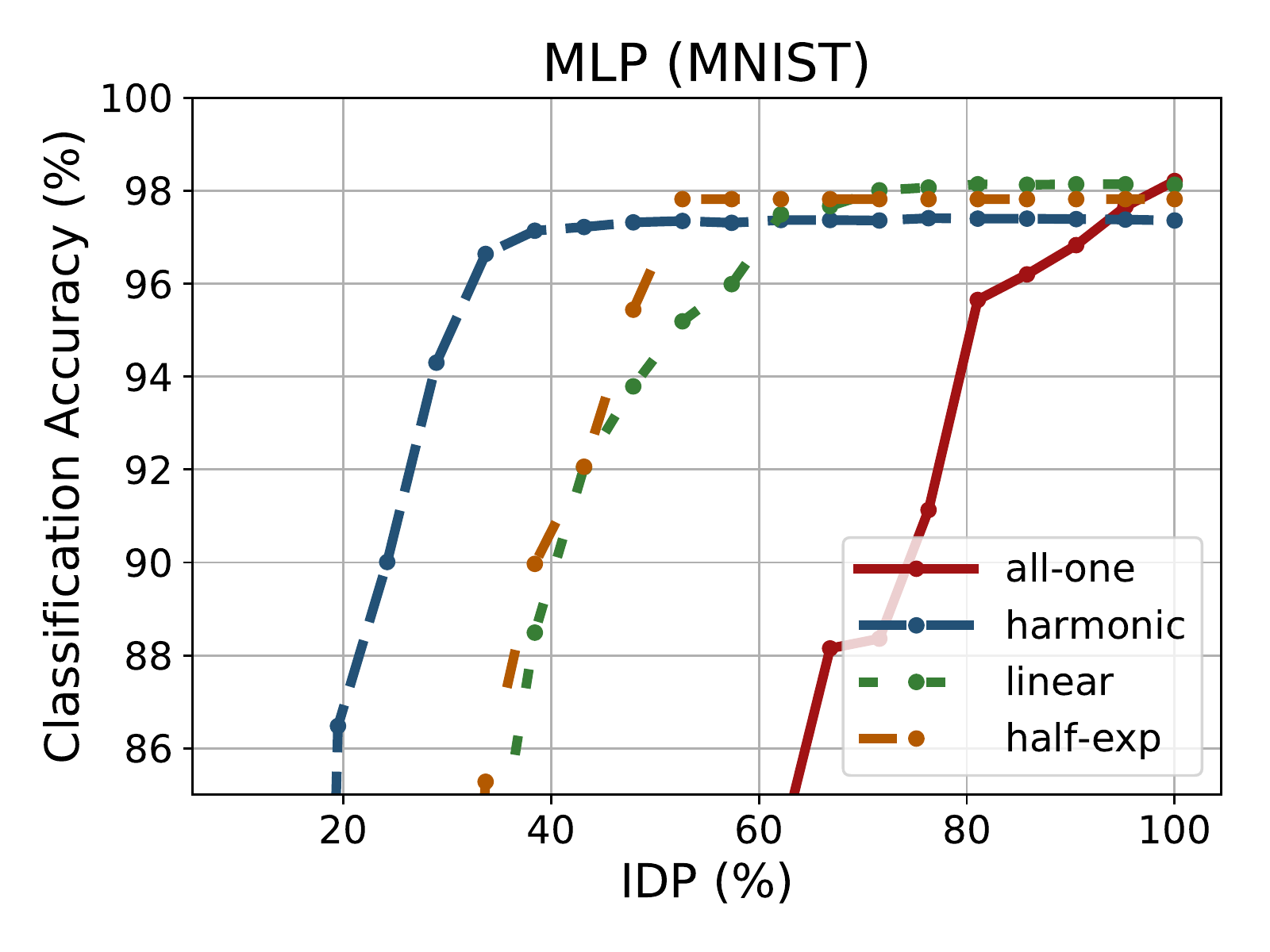}
    \caption{A comparison of the classification accuracy of the MLP structure in Figure~\ref{fig:networks} (a), trained using the four profiles (shown in Figure~\ref{fig:coef-functions}) for the MNIST dataset. The x-axis shows IDP (\%), which is the percentage of components used in the IDP layer during forward propagation. The all-one profile is equivalent to the standard network (without channel coefficients).}
    \label{fig:channel-coef}
\end{figure}

The choice of profile has a large impact on the IDP range of a trained model. Figure~\ref{fig:channel-coef} shows the effect of different profiles for the MLP. The MLP is used for this comparison since the smaller model size magnifies the differences in behavior of networks trained with the various profiles. Using the all-one profile (shown in red) is equivalent to training a standard network without IDP. We see that this all-one model achieves the highest accuracy result when the dot product is complete ($100\%$ IDP), but falls off quickly with poorer performance as IDP is decreased. This shows that the standard model does not support a large dynamic computation range. By equally weighting the contribution of each channel, the quality of the network deteriorates more quickly as fewer of the channels are used. 

The other profiles weight the early channels more and the latter channels less. This allows a network to deteriorate more gradually, as the channels with the smaller contribution are unused in the IDP. As an example, the MLP model trained with the linear profile is able to maintain higher accuracy from $60$-$100\%$ IDP. At $100\%$ IDP, the linear profile model still achieves the same accuracy as the standard model. We will compare the standard network (all-one profile) to linear profile for the remaining analysis in the paper.

Compared to the linear profile, the harmonic profile places a much higher weight on the beginning subset of channels. This translates to a larger dynamic computation range, from $30$-$100\%$ IDP, for a model trained with the harmonic profile. However, this model performs about $1\%$ worse than the standard model when all channels are used ($100\%$ IDP). Fortunately, as described in Section~\ref{sec:multi-eval}, we can generally use multiple profiles in a single network to mitigate this problem of lowered accuracy in high IDP percentage regions.

\subsection{Single-Profile Incomplete Neural Networks}
\label{sec:coef-eval}
\begin{figure*}
    \begin{subfigure}{.5\textwidth}
      \centering
      \includegraphics[width=.9\linewidth]{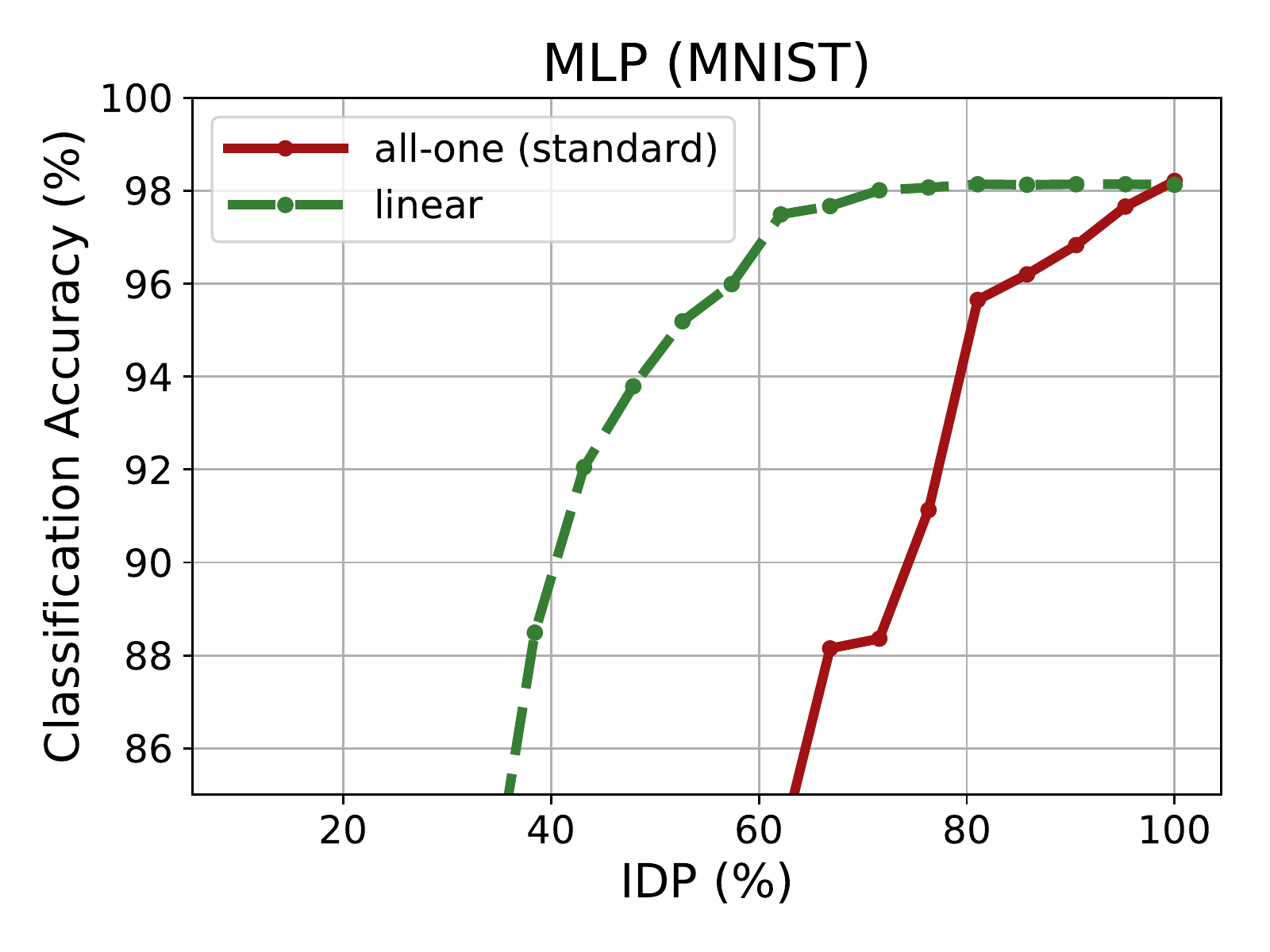}
    \end{subfigure}%
    \begin{subfigure}{.5\textwidth}
      \centering
      \includegraphics[width=.9\linewidth]{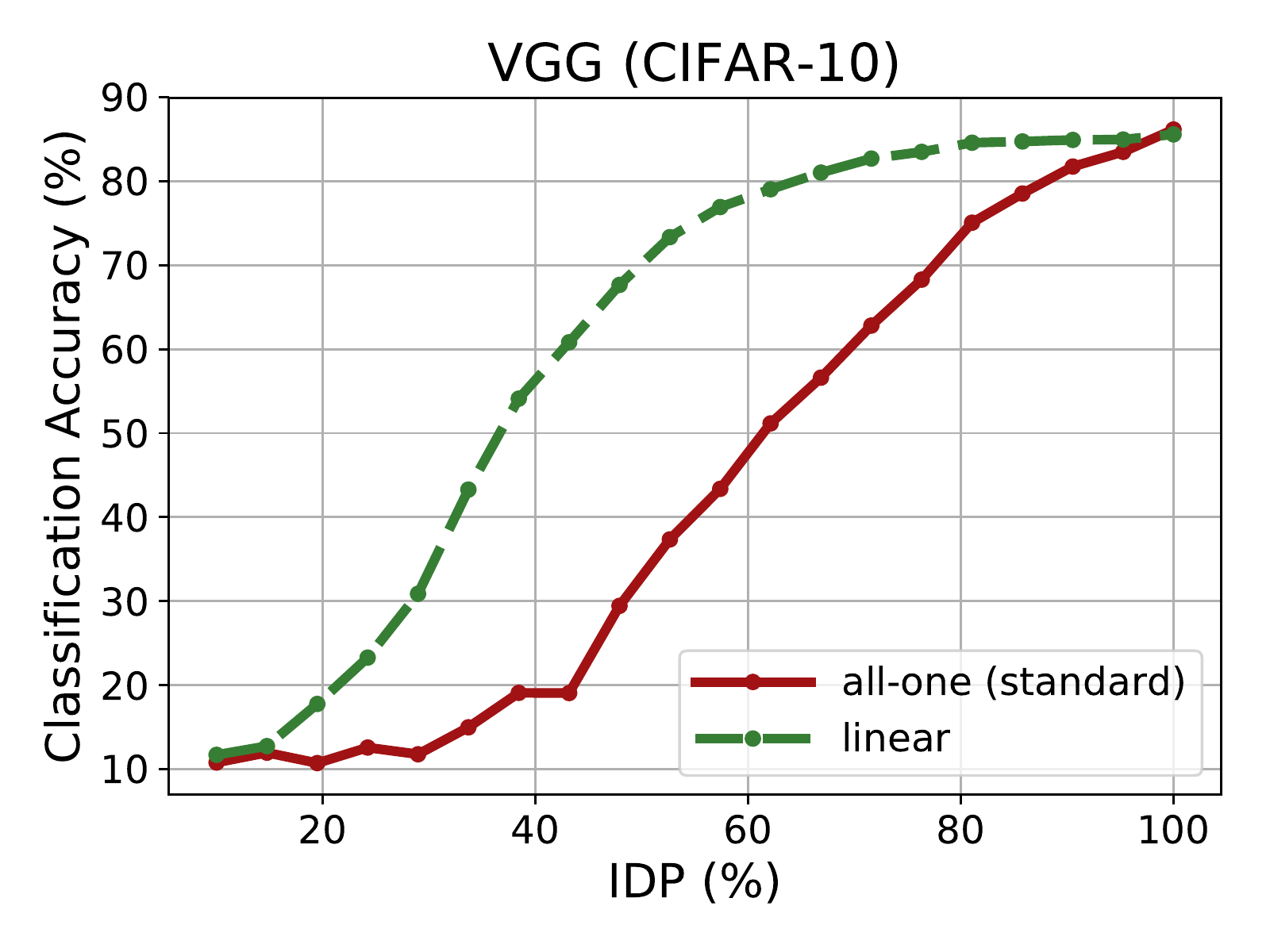}
    \end{subfigure}
    \begin{subfigure}{.5\textwidth}
      \centering
      \includegraphics[width=.9\linewidth]{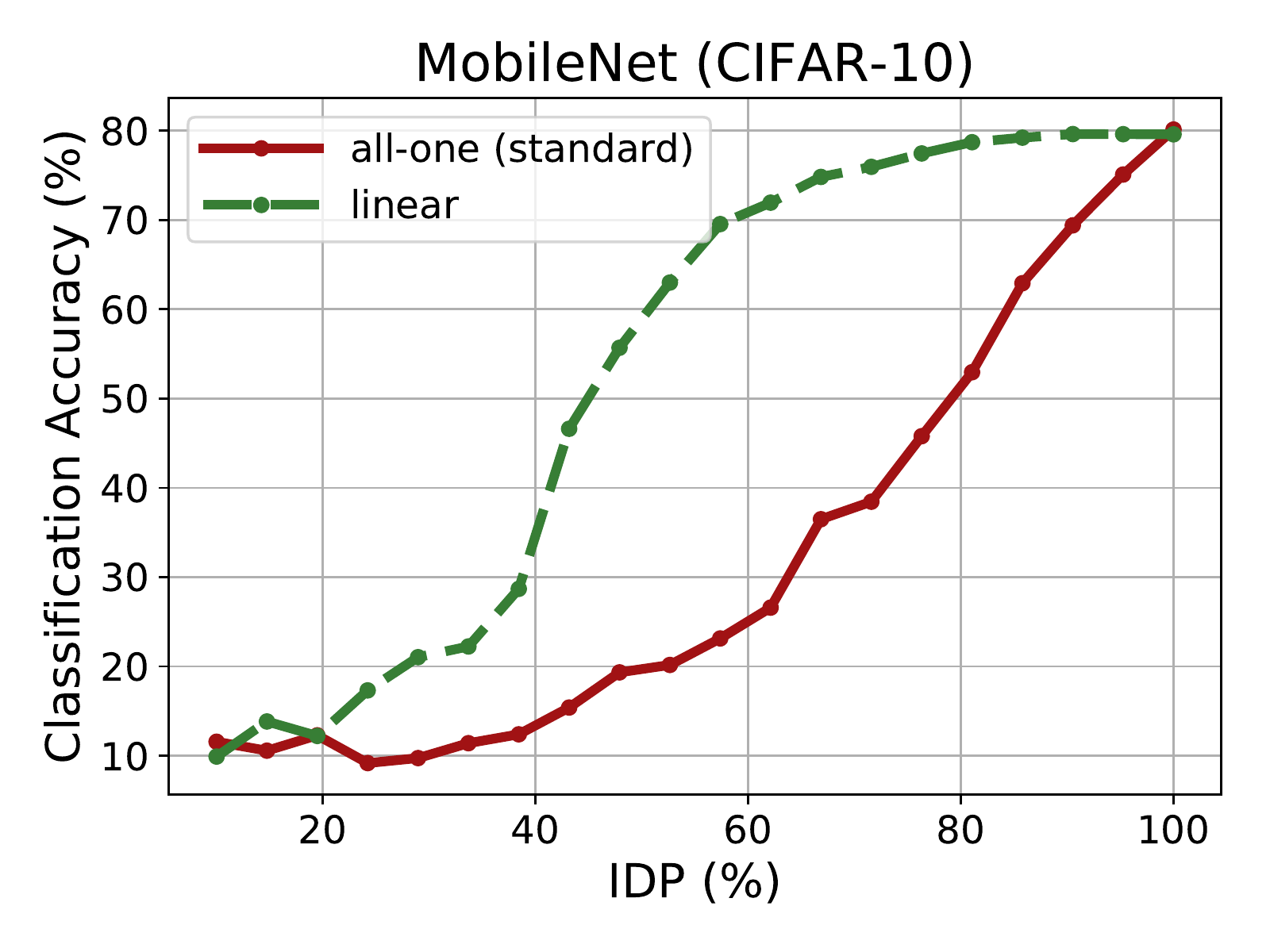}
    \end{subfigure}%
    \begin{subfigure}{.5\textwidth}
      \centering
      \includegraphics[width=.9\linewidth]{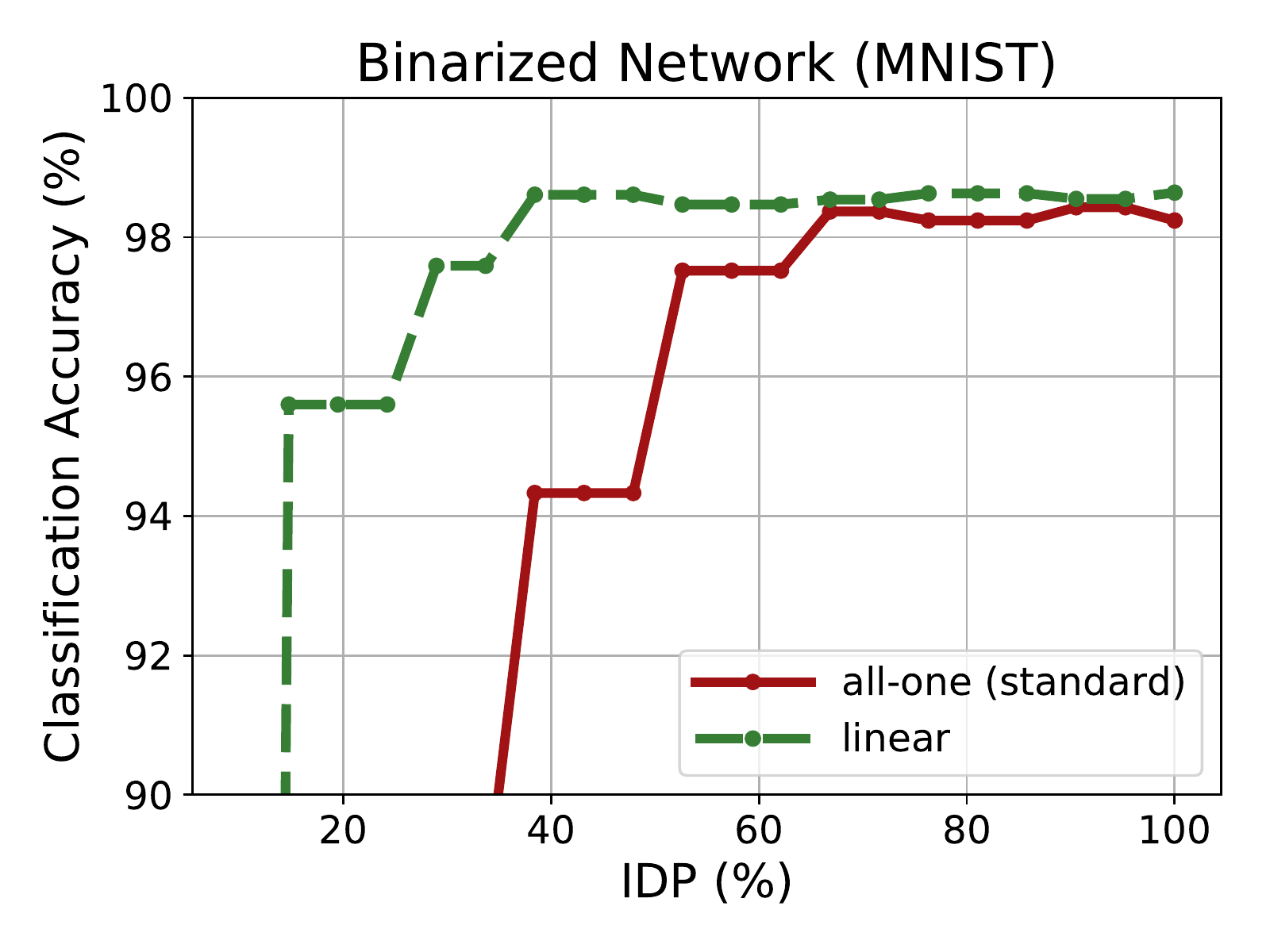}
    \end{subfigure}

    \caption{A comparison of the performance of dynamic scaling with IDP for standard networks (trained without a profile) to incomplete networks (trained with the linear profile), for each network structure described in Section~\ref{sec:networks}. Top-left is MLP. Top-right is VGG-16. Bottom-left is MobileNet and bottom-right is Binarized Network.}
    \label{fig:idp-comparison}
\end{figure*}

In this section, we evaluate the performance of the networks presented in Figure~\ref{fig:networks} trained with the linear profile compared to a standard network. Figure~\ref{fig:idp-comparison} compares the dynamic scaling of incomplete dot products for standard networks to incomplete networks over four different networks: MLP, VGG-16, MobileNet and Binarized Networks.

For each network structure, we observe that using the linear profile allows the network to achieve a larger dynamic computation range compared to the standard network. For the VGG-16 (CIFAR-10) model, at $50\%$ IDP, the model with the linear profile achieves an accuracy of $70\%$ as opposed to $35\%$ by all-one (standard) model. Additionally, for each network the linear IDP network is able to achieve the same accuracy as the standard network when all channels are used.

\subsection{Multiple-Profile Incomplete Neural Networks}
\label{sec:multi-eval}
\begin{figure*}
    \begin{subfigure}{.5\textwidth}
      \centering
      \includegraphics[width=.9\linewidth]{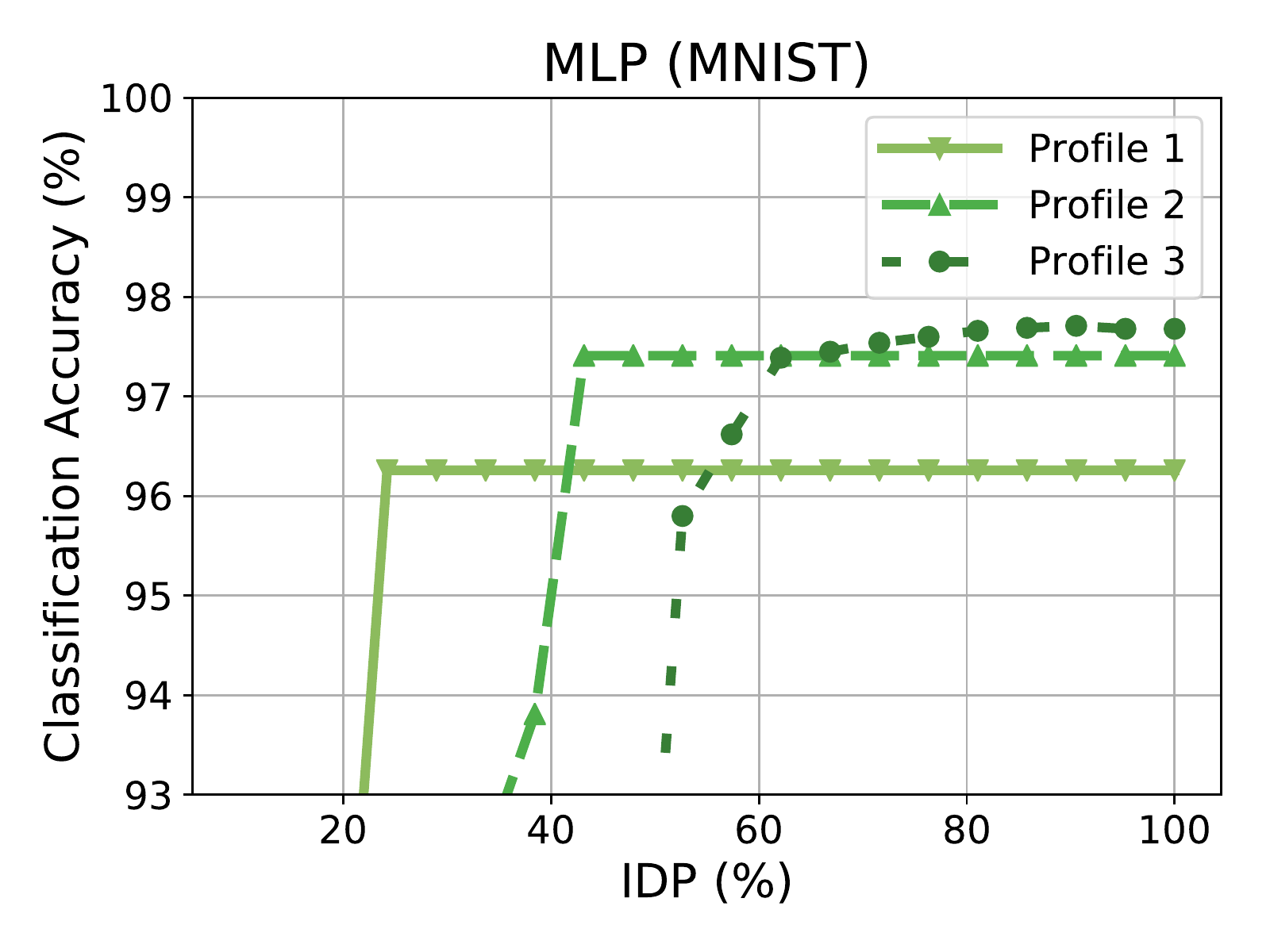}
    \end{subfigure}%
    \begin{subfigure}{.5\textwidth}
      \centering
      \includegraphics[width=.9\linewidth]{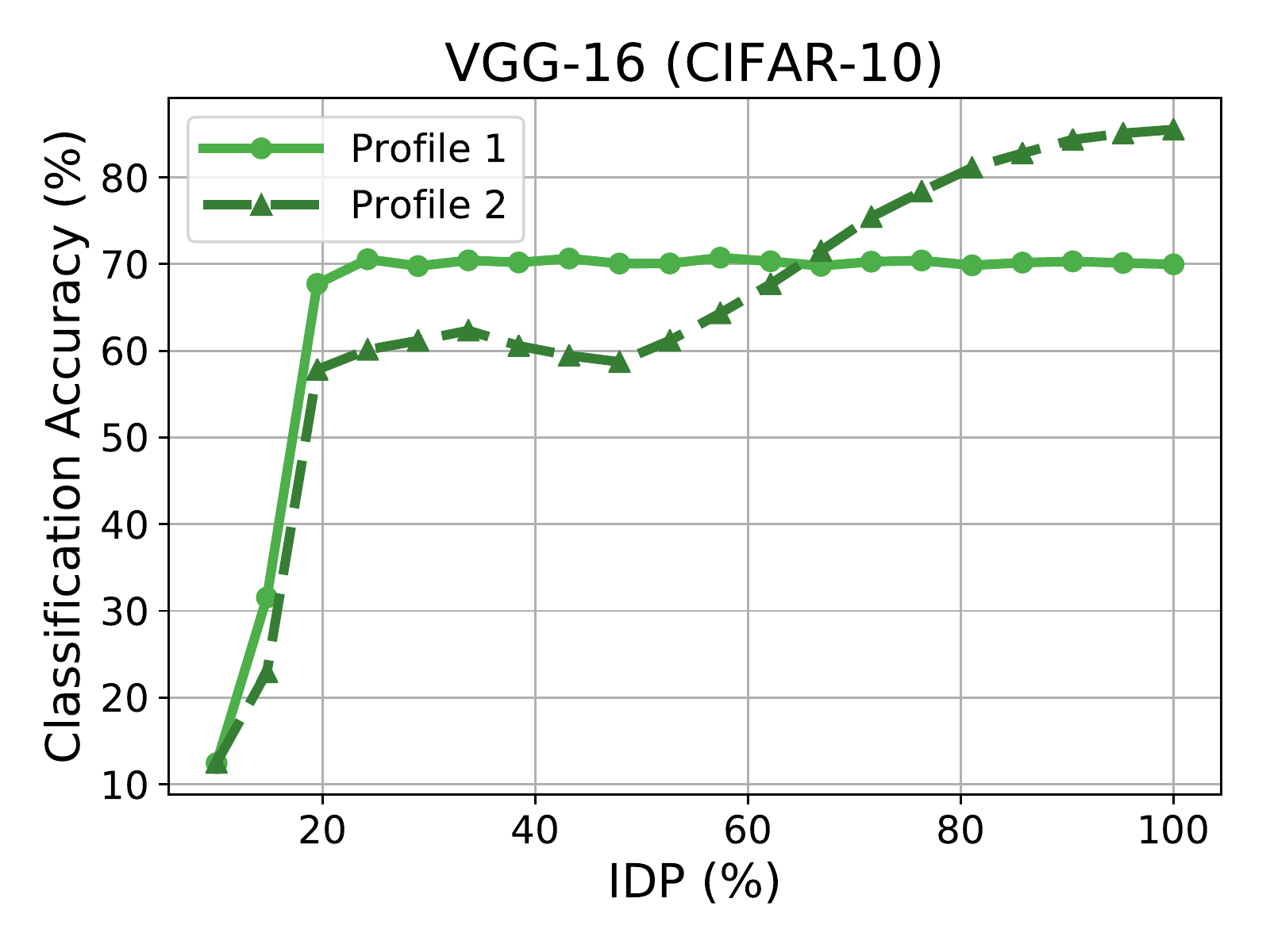}
    \end{subfigure}

    \caption{A comparison of performance of dynamic IDP scaling under three profiles for MLP (left) and two profiles for VGG-16 (right). All profiles for a given model share the same network weights in the IDP layers. Both networks are trained using the linear coefficients.}
    \label{fig:multi-idp-comparison}
\end{figure*}

In this section, we explore the performance of multiple-profile incomplete neural networks, as described in Section~\ref{sec:multi-profile}. Figure~\ref{fig:multi-idp-comparison} shows how each profile in a multiple-profile incomplete network scales across the IDP percentage range for the MLP and VGG-16 networks. For the MLP, a three profile scheme is used, where the first profile is trained for the $0$-$20\%$ range, the second profile for the $20$-$40\%$ range, and the third profile for the $40$-$100\%$ range. For VGG-16, a two profile scheme is used, where the first profile is trained for the $0$-$30\%$ range and the second profile for the $30$-$100\%$ range. The profiles are trained incrementally, starting with the first profile. Each profile only learns the weights in its specified range~(\eg,~$30$-$100\%$), but utilizes the weights learned by earlier profiles~(\eg,~$0$-$30\%$). In this way, the later profiles can use weights learned from the earlier profiles without affecting the performance of the earlier profiles. Training the network in this multi-stage fashion enables a single set of weights to support multiple profiles.

For the first profile of the VGG-16 model, we observe that the accuracy does not improve when an IDP of greater than $30\%$ is used. Since $30\%$ IDP is the maximum for the profile, it does not learn the channels in the $30$-$100\%$ IDP range, and therefore cannot use the higher ranges during inference. The second profile is able to achieve a higher final accuracy than the first profile but performs worse in the lower part of the IDP range. The two profiles are able to achieve a higher accuracy across the entire IDP range compared to the single profile network shown in Figure~\ref{fig:idp-comparison}.

By training the profiles in a multi-stage fashion, the first profile is restricted to learn a classifier using only the first $30\%$ of channels. This improves the accuracy of the model in the lower IDP regions compared to the single-profile case. For instance, profile 1 achieves a $70\%$ classification accuracy at $30\%$ IDP compared to only $30\%$ accuracy at $30\%$ IDP in the single profile case. While profile 2 does not update the channels learned by profile 1, it still utilizes them during training. Note that profile 2 can still achieve similar performance to the single-profile model in the $80$-$100\%$ IDP region.

For the MLP model, we observe similar trends, but applied to a three profile case. As more profiles are added, we see that the final profile (profile 3) is still able to achieve a high final accuracy, even though the beginning $40\%$ of input channels in the IDP layer are shared between all three profiles.

% In training multiple profiles, we anticipate its weights will be shared by later profiles. As such we leave some positive weights for high percentage range.

% \subsection{Comparing Network Sizes over a Computation Range}
% Large IDP network can scale across the range with similar performance to smaller standard network as shown in Figure~\ref{fig:idp-comparison}.

%\subsection{Comparing Binary and Ternary Activation}
%For the binarized network model, we compare binary and ternary activation for MNIST. Justify our choice of coefficient function...

%\begin{figure}
%    \centering
%    \includegraphics[width=\linewidth]{activ_comparison_mnist}
%    \caption{The classification accuracy for MNIST of the edge-cloud network structure trained using either binary activation or ternary activation with $\varepsilon$ of $0.5$, $1$, and $2$. In this plot, all networks were trained using the harmonic channel coefficient function. We observe that the binary activation network drops off more quickly in accuracy as fewer components are used than the corresponding ternary networks.}
%    \label{fig:mnist-activ-comp}
%\end{figure}

% Describe how the input and output layers need to added for each additional profile and this additional cost is low

\section{Related Work}
\label{sec:related}
In this section, we first compare IDP to methods that are similar in style but have the objective of preventing overfitting instead of providing a dynamic mechanism to scale over a computation range. Dropout~\cite{srivastava2014dropout} is a popular technique for dealing with overfitting by using only a subset of features in the model for any given training batch. At test time, dropout uses the full network whereas IDP allows the network to dynamically adjust its computation by using only a subset of channels. DropConnect~\cite{wan2013regularization} is similar to Dropout, but instead of dropping out the output channels, it drops out the network weights. This approach is similar to IDP, as IDP is also a mechanism for removing channels in order to support dynamic computation scaling. However, IDP adds a profile to directly order the contribution of the channels and does not randomly drop them during training.  DeCov~\cite{cogswell2015reducing} is a more recent technique for reducing overfitting which directly penalizes redundant features in a layer during training. At a high level, this work shares a similar goal with  multiple-profile IDP, by aiming to create a set of non-redundant channels that generalizes well given the restricted computation range of a single profile.

There is a growing body of work on CNN models that have a smaller memory footprint~\cite{mcdanel2017embedded} and are more power efficient. One of the driving innovations is depthwise separable convolution~\cite{chollet2016xception}, which decouples a convolutional layer into a depthwise and pointwise layer, as shown in Figure~\ref{fig:incomplete_separable}. This approach is a generalization of the inception module first introduced in GoogLeNet~\cite{szegedy2015going, szegedy2016rethinking, szegedy2017inception}. MobileNet~\cite{howard2017mobilenets} utilized depthwise separable convolutions with the objective of performing state of the art CNN inference on mobile devices. ShuffleNet~\cite{zhang2017shufflenet} extends the concept of depthwise separable convolution and divides the input into groups to further improve the efficiency of the model. Structured Sparsity Learning (SSL)~\cite{wen2016learning} constrains the structure of the learned filters to reduce the overall complexity of the model. IDP is orthogonal to these approaches and can be used in conjunction with them, as we show by incorporating IDP in MobileNet.

%AlexNet\cite{krizhevsky2012imagenet} introduces an efficient implementation of distributed training over multiple GPUs by dividing the input into groups. 
%GoogLeNet\cite{szegedy2015going} adds an inception module to lower the complexity of stacking deep convolutional layers. SqueezeNet\cite{iandola2016squeezenet} adds squeeze and expand layers to significantly reduce the number of parameters of a neural network. 
%ResNet\cite{he2016deep} adds a bottleneck module with 1x1 convolution layers for reducing and restoring dimensions to further reduce the complexity of deep neural networks. 

% \section{Discussion}

% \subsection{Speedup / Pros / Cons of IDP}
% \subsection{Complexity of IDP}
% Comparison of inference computation using complete dot  product  (CDP)  in  standard  networks  and  incomplete  dot product (IDP) presented in this paper for a convolutional layer.In standard CNN inference (before), for each filter, all channels are  used  in  the  dot  product  computation  (CDP), to compute each output channel. Under, for example,50\% IDP (now), half of the filters (such as F1) are used, each using only 50\% of the filter channels to compute each output channel. The remaining50\% of  filters  (such  as Fn),  are  unused  since  their  channels will  not  be  utilized  in  the  next  layer.  This  leads  to  a75\% reduction in computation for50\% IDP.

% $p^2$ reduction in computation.

% \subsection{Binary Neural Network Storage and Computation Reduction}

\section{Future Work}
A profile targeted for a specific application can use any number of channels in the given IDP range. The current implementation approach, as described in Section~\ref{sec:multi-profile} of this paper, aims to have the profiles share the same coefficients on overlapping channels. For instance, in Figure~\ref{fig:training}, the first half of the profile 2 coefficients are the profile 1 coefficients. To this end, we train the weights incrementally starting with the innermost profile and extending to the outermost profile. In the future, we want to study a more general setting, where this coefficient sharing constraint could be relaxed by jointly training both the network weights and the profile coefficients.

%Instead of enforcing this restriction, it may be possible for the coefficients of each profile to be independent while still sharing the same network weights. In this formulation, we could also remove the constraint that coefficients are monotonically non-increasing and non-negative. This may allow the coefficients of profiles to be adjusted on an individual profile basis in order to improve classification accuracy. For example, a channel may acquire a large coefficient for good performance in profile 1, and therefore require a large coefficient. However in profile 2, the channel may acquire a different  coefficient similar to those of other channels in an expanded IDP range, making it less important.

%Additionally, rather than using a profile to set the channel coefficients, such as those shown in Figure~\ref{fig:coef-functions}, the network may achieve higher accuracy if the coefficients were learned directly from data. In the case of multiple profiles, all of the profile coefficient sets, one per profile, could be jointly trained together with the same shared network weights. This would allow a profile with fewer components to better adjust the coefficients to emphasize the discriminative features in other channels. We hope that this joint training approach would remove the limitation of shared coefficients imposed by methods such as incremental multi-stage training described in Section~\ref{sec:multi-profile}.

\section{Conclusion}
This paper proposes incomplete dot product (IDP), a novel way to dynamically adjust the inference costs based on the current computation budget in conserving battery power or reducing application latency of the device. IDP enables this by introducing profiles of monotonically non-increasing channel coefficients to the layers of CNNs. This allows a single network to scale the amount of computation at inference time by adjusting the number of channels to use (IDP percentage). As illustrated in Figure~\ref{fig:savings-comparison}, IDP has two sources for its computation saving, and their effects are multiplicative. For example, $50\%$ IDP will lead to reduce computation by $75\%$.

Additionally, we can improve the flexibility and effectiveness of IDP at inference time by introducing multiple profiles. Each profile is trained to target a specific IDP range. At inference time, the current profile is chosen by the target IDP range, which is selected by the application or according to a power management policy. By using multiple profiles, we are able to train a network which can run over a wide computation scaling range while maintaining a high accuracy (see Figure~\ref{fig:multi-idp-comparison}).

To the best of our knowledge, the dynamic adaptation approach of IDP as well as the notion of multiple profiles and network training for these profiles are novel. As CNNs are increasingly running on devices ranging from smartphones to IoT devices, we believe methods that provide dynamic scaling such as IDP become more important. We hope that this paper can inspire further work in dynamic neural network reconfiguration, including new IDP designs, training and associated methodologies.

\section{Acknowledgements}
This work is supported in part by gifts from the Intel Corporation
and in part by the Naval Supply Systems Command
award under the Naval Postgraduate School Agreement No. N00244-16-1-0018.

% To allow flexible and effective IDP configurations at the inference time, we have presented a novel training methodology.  We show how the distribution of training batches for different IDP configurations can reflect intended dynamics during inference. At inference time, DNNs trained with IDP can scale down by a factor of 2x in communication while maintaining similar accuracy to a standard network with the same number of filters. By expanding the output channels in the cloud, IDP is able to further improve the accuracy without introducing additional communication between the device and cloud. The source code for IDP is written with Chainer~\cite{chainer_learningsys2015} and will be provided when the paper is accepted.

%To validate IDP performance, we have considered two standard benchmark tasks, MNIST and CIFAR-10. We demonstrate that IDP can allow aggressive implementation on devices with extremely limited resources while achieving competitive classification accuracy. For example, for CIFAR-10, we can achieve xx\%? or KY\%? accuracy while using only xKB or yKB communication, as opposed xxKB or yyKB, when the entire MNIST or CIFAR input image, respectively, needs to transmitted to the cloud as in typical offloading.

%\section*{Acknowledgment}

\bibliographystyle{IEEEtran}
\bibliography{IEEEabrv,bibliography}

\begin{thebibliography}{10}
\providecommand{\url}[1]{#1}
\csname url@samestyle\endcsname
\providecommand{\newblock}{\relax}
\providecommand{\bibinfo}[2]{#2}
\providecommand{\BIBentrySTDinterwordspacing}{\spaceskip=0pt\relax}
\providecommand{\BIBentryALTinterwordstretchfactor}{4}
\providecommand{\BIBentryALTinterwordspacing}{\spaceskip=\fontdimen2\font plus
\BIBentryALTinterwordstretchfactor\fontdimen3\font minus
  \fontdimen4\font\relax}
\providecommand{\BIBforeignlanguage}[2]{{%
\expandafter\ifx\csname l@#1\endcsname\relax
\typeout{** WARNING: IEEEtran.bst: No hyphenation pattern has been}%
\typeout{** loaded for the language `#1'. Using the pattern for}%
\typeout{** the default language instead.}%
\else
\language=\csname l@#1\endcsname
\fi
#2}}
\providecommand{\BIBdecl}{\relax}
\BIBdecl

\bibitem{McMahanMRA16}
\BIBentryALTinterwordspacing
H.~B. McMahan, E.~Moore, D.~Ramage, and B.~A. y~Arcas, ``Federated learning of
  deep networks using model averaging,'' \emph{CoRR}, vol. abs/1602.05629,
  2016. [Online]. Available: \url{http://arxiv.org/abs/1602.05629}
\BIBentrySTDinterwordspacing

\bibitem{howard2017mobilenets}
A.~G. Howard, M.~Zhu, B.~Chen, D.~Kalenichenko, W.~Wang, T.~Weyand,
  M.~Andreetto, and H.~Adam, ``Mobilenets: Efficient convolutional neural
  networks for mobile vision applications,'' \emph{arXiv preprint
  arXiv:1704.04861}, 2017.

\bibitem{simonyan2014very}
K.~Simonyan and A.~Zisserman, ``Very deep convolutional networks for
  large-scale image recognition,'' \emph{arXiv preprint arXiv:1409.1556}, 2014.

\bibitem{lecun1998mnist}
Y.~LeCun, C.~Cortes, and C.~J. Burges, ``The mnist database of handwritten
  digits,'' 1998.

\bibitem{krizhevsky2014cifar}
A.~Krizhevsky, V.~Nair, and G.~Hinton, ``The cifar-10 dataset,'' 2014.

\bibitem{teerapittayanon2017distributed}
S.~Teerapittayanon, B.~McDanel, and H.~Kung, ``Distributed deep neural networks
  over the cloud, the edge and end devices,'' in \emph{37th International
  Conference on Distributed Computing Systems (ICDCS 2017)}, 2017.

\bibitem{courbariaux2016binarized}
M.~Courbariaux, I.~Hubara, D.~Soudry, R.~El-Yaniv, and Y.~Bengio, ``Binarized
  neural networks: Training deep neural networks with weights and activations
  constrained to+ 1 or-1,'' \emph{arXiv preprint arXiv:1602.02830}, 2016.

\bibitem{srivastava2014dropout}
N.~Srivastava, G.~E. Hinton, A.~Krizhevsky, I.~Sutskever, and R.~Salakhutdinov,
  ``Dropout: a simple way to prevent neural networks from overfitting.''
  \emph{Journal of Machine Learning Research}, vol.~15, no.~1, pp. 1929--1958,
  2014.

\bibitem{wan2013regularization}
L.~Wan, M.~Zeiler, S.~Zhang, Y.~L. Cun, and R.~Fergus, ``Regularization of
  neural networks using dropconnect,'' in \emph{Proceedings of the 30th
  International Conference on Machine Learning (ICML-13)}, 2013, pp.
  1058--1066.

\bibitem{cogswell2015reducing}
M.~Cogswell, F.~Ahmed, R.~Girshick, L.~Zitnick, and D.~Batra, ``Reducing
  overfitting in deep networks by decorrelating representations,'' \emph{arXiv
  preprint arXiv:1511.06068}, 2015.

\bibitem{mcdanel2017embedded}
B.~McDanel, S.~Teerapittayanon, and H.~Kung, ``Embedded binarized neural
  networks,'' \emph{arXiv preprint arXiv:1709.02260}, 2017.

\bibitem{chollet2016xception}
F.~Chollet, ``Xception: Deep learning with depthwise separable convolutions,''
  \emph{arXiv preprint arXiv:1610.02357}, 2016.

\bibitem{szegedy2015going}
C.~Szegedy, W.~Liu, Y.~Jia, P.~Sermanet, S.~Reed, D.~Anguelov, D.~Erhan,
  V.~Vanhoucke, and A.~Rabinovich, ``Going deeper with convolutions,'' in
  \emph{Proceedings of the IEEE conference on computer vision and pattern
  recognition}, 2015, pp. 1--9.

\bibitem{szegedy2016rethinking}
C.~Szegedy, V.~Vanhoucke, S.~Ioffe, J.~Shlens, and Z.~Wojna, ``Rethinking the
  inception architecture for computer vision,'' in \emph{Proceedings of the
  IEEE Conference on Computer Vision and Pattern Recognition}, 2016, pp.
  2818--2826.

\bibitem{szegedy2017inception}
C.~Szegedy, S.~Ioffe, V.~Vanhoucke, and A.~A. Alemi, ``Inception-v4,
  inception-resnet and the impact of residual connections on learning.'' in
  \emph{AAAI}, 2017, pp. 4278--4284.

\bibitem{zhang2017shufflenet}
X.~Zhang, X.~Zhou, M.~Lin, and J.~Sun, ``Shufflenet: An extremely efficient
  convolutional neural network for mobile devices,'' \emph{arXiv preprint
  arXiv:1707.01083}, 2017.

\bibitem{wen2016learning}
W.~Wen, C.~Wu, Y.~Wang, Y.~Chen, and H.~Li, ``Learning structured sparsity in
  deep neural networks,'' in \emph{Advances in Neural Information Processing
  Systems}, 2016, pp. 2074--2082.

\end{thebibliography}

\end{document}